%% file: main.tex
\pdfoutput=1
\documentclass[11pt]{article}
\usepackage{ACL2023}

\usepackage{times}
\usepackage{latexsym}
\usepackage[T1]{fontenc}
\usepackage[utf8]{inputenc}
\usepackage{microtype}
\usepackage{inconsolata}

\usepackage{graphicx}
\usepackage{times}
\usepackage{fancyhdr,graphicx,amsmath,amssymb}
\usepackage[ruled,vlined]{algorithm2e}
\usepackage{multirow}
\usepackage{amsfonts}

\title{A Dialogue Game for Eliciting Balanced Collaboration}

\author{Isidora Jeknić \\
  Saarland University \\
  Saarbrücken, Germany\\
  \footnotesize{{\tt jeknic@lst.uni-saarland.de}
  }
  \\\And
  David Schlangen \\
  University of Potsdam \\
  Potsdam, Germany\\
  \footnotesize{{\tt david.schlangen@uni-potsdam.de}}
  \\\And
  Alexander Koller \\
  Saarland University \\
  Saarbrücken, Germany\\
  \footnotesize{{\tt koller@coli.uni-saarland.de}}
  }

\begin{document}
\maketitle
\begin{abstract}
	Collaboration is an integral part of human dialogue. 
  Typical task-oriented dialogue games assign asymmetric roles
  to the participants, which limits their ability to elicit
  naturalistic role-taking in collaboration and its negotiation.
	We present a novel and simple online setup that favors balanced collaboration: a two-player 2D object placement game in which the players must negotiate the goal state themselves.
  We show empirically that human players exhibit a variety of role distributions, and that balanced collaboration improves task performance.
  We also present an LLM-based baseline agent which demonstrates that automatic playing of our game is an interesting challenge for artificial systems.
\end{abstract}

\input{introduction}
\input{background}
\input{game}
\input{strategies}
\input{modeling}
\input{conclusion}

\section*{Limitations}
The dominance score is a useful operationalization of the collaborative imbalance between players, but it is an approximation that does not actually take the content of the players' chat messages into account. It might be interesting to refine this measure in the future, e.g.\ by having the messages evaluated by an LLM. Nevertheless, the post-hoc manual analysis of the games in the four strategies indicates that the dominance score captures differences in collaboration strategy well.

Additionally, the LLM agent presented in Section~\ref{sec:modeling} is a relatively simple baseline. It is conceivable that a more intricate LLM model would close the gap to human performance, at least to the Leader strategy. We leave the exploration of such models, and of more intricate versions of our game that would remain challenging for them, for future research.

\section*{Ethics Statement}
We do not see any particular ethics challenges with the research reported here.

\bibliography{bibliography,custom}
\bibliographystyle{acl_natbib}

\newpage
\appendix
\input{appendix}

\end{document}

%% file: introduction.tex

\section{Introduction}

Language use is a highly collaborative process that involves constant negotiation and cooperation between interlocutors, with the ultimate goal of facilitating mutual understanding \cite{clark_1986_referring, clark_1996_using, grice_1989_studies}.
An improved understanding of these negotiation processes would benefit the development of future systems for effective human-AI cooperation and go beyond today's rigid division of roles between dialogue systems and users \citep{dafoe2020open, dafoe2021cooperative}.

Collaborative dialogue is often studied through dialogue games involving reconstruction, where one player has information about a target configuration and guides the other player towards it \cite{clark_1986_referring, zarrie_2016_pentoref, kim-etal-2019-codraw, lachmy2022draw}. This approach places the players in fixed roles (instruction giver/follower), which is in contrast to the fluid and implicit negotiation of these roles in naturally occurring collaborative dialogue. This limits the ability of such games to elicit negotiation about collaborative roles.

In this paper, we directly address this issue by introducing a collaborative game designed to elicit dialogues with more flexible role-taking -- a 2D object placement game in which the target configuration is not predetermined, but must be negotiated by the players. The players use online chat to jointly decide how to arrange movable objects, without seeing each other's boards.

We observe that players indeed exhibit a variety of collaboration strategies in this dialogue game, further illustrated by a metric we define, the dominance score, representing the degree to which one player controls the gameplay. Only a minority of player dyads choose an asymmetric strategy in which one player always dominates; and this strategy is also associated with systematically lower scores than more balanced strategies. Finally, we describe a baseline computational agent for this game. It achieves a significantly lower average score than a human player using a limited collaboration strategy, indicating that natural and effective collaboration in balanced games like ours is an interesting avenue for future research.\footnote{Our code and data are available at: \url{https://github.com/coli-saar/placement-game}.}

\begin{figure*}[t]
    \centering
    \includegraphics[width=1\linewidth]{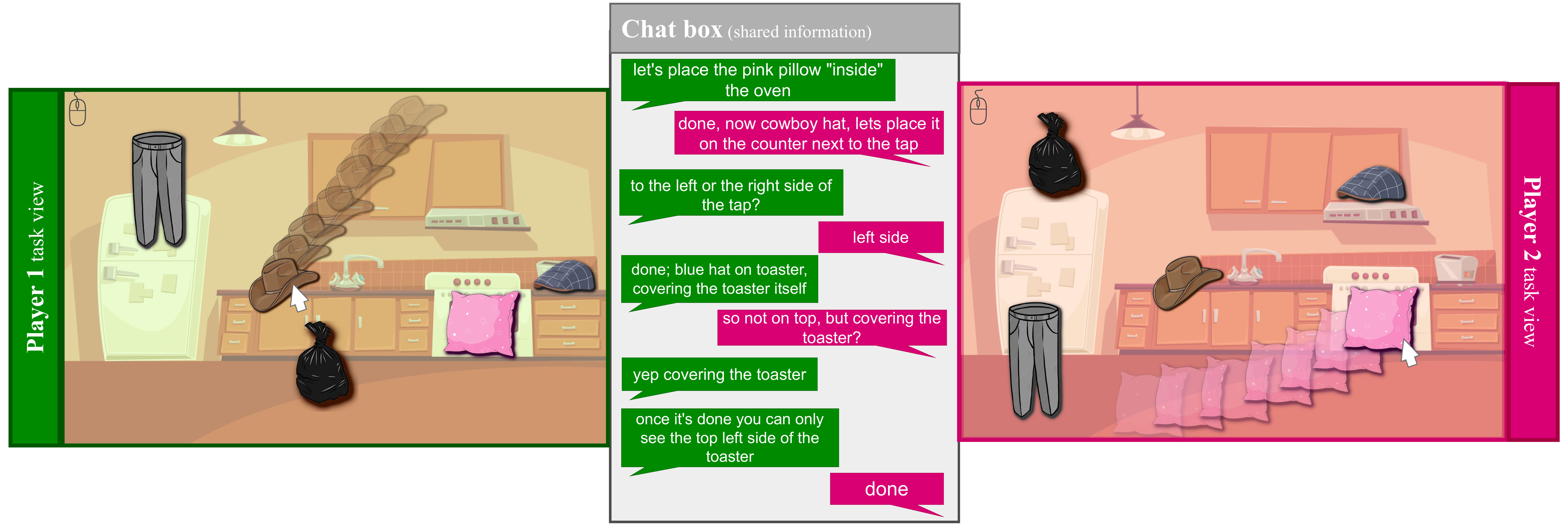}
    \caption{A reconstructed task view of both players illustrating the shared information (middle, chat box) and information only available to each respective player (left and right). Additionally, illustrates an instance of the back and forth strategy (as described in Section \ref{sec:strats}).}
    \label{fig:view}
\end{figure*}

%% file: background.tex

\section{Background}

\textbf{Collaboration in dialogue.}	In situated dialogue, common ground and shared context are paramount to avoiding misunderstandings \cite{clark_1996_using,brownschmidt_2018_perspectivetaking}. Interlocutors commonly engage in a collaborative effort, i.e., negotiation, to establish these commonalities, and, ultimately, a joint purpose \cite{clark_1996_using}, frequently through a coordinated referencing approach \citep{clark_1986_referring}. In order to arrive at a unified goal, the interlocutors must work together and coordinate their actions over time \cite{pickering_2013_an}. This ongoing coordination process can lead to the acquisition of new knowledge, including how to coordinate better \citep{schlangen_2023_what}. 

\noindent \textbf{Collaborative games.} There have been many reconstruction game environments developed for the purpose of studying collaboration and negotiation (e.g., \citet{zarrie_2016_pentoref, kim-etal-2019-codraw, pacella2022understanding, narayanchen_2019_collaborative}; for overview see \citet{suglia2024visually}). However, all previously cited environments assume a predetermined target state to which one player must guide the other. This inherently places the participants on different levels dependent on the role they are given (instructor vs. follower), determined by the information they are given. While this is a legitimate collaboration strategy, these environments are too restrictive insofar as they do not allow more balanced approaches, which are necessary for a holistic study of collaboration \cite{schlangen2018meetup}.

\noindent \textbf{Human-computer collaboration.}		Here we refer to all collaborative situations in which ``agents may be able to achieve joint gains or avoid joint losses'' \citep[8]{dafoe2020open}. In the field of human-computer dialogue systems, the most frequent such agents are instruction-giving \citep{koller2010first, kohn2020generating,sadler2024learning, janarthanam2010learning, narayanchen_2019_collaborative, zarrie_2016_pentoref}, or instruction-following  \citep{hill2020human, chan2019actrce}. While they do involve a level of first-hand human-computer interaction and dialogue necessary for completing a given task, both cases are characterized by a built-in asymmetry, analogous to the aforementioned reconstruction games. In order to ensure successful and robust human-computer cooperation, and facilitate trust, it is integral for inherently collaborative systems (e.g., assistants) to be able to handle balanced collaboration, as well \cite{dafoe2020open}.

%% file: game.tex

\section{Collaborative object-placement game}

We developed a collaborative, 2D object placement game that can be played by two players over the Internet. In each round, the two players see an identical, static background, upon which movable objects have been placed in random positions that are different for the two players (see Figure \ref{fig:view}). The goal of the game is for the players to place each object in the same position by dragging it with the mouse. Players cannot see each other's scene; they can only communicate through a chat window.

Each pair of players played two rounds of the game together, with a kitchen background in the first round and a living room in the second (see Appendix \ref{app:design} for more images). This allowed us to study how their collaboration strategies evolved as they became more familiar with each other.

We make the game available online by integrating it into Slurk \citep{schlangen2018slurk, gotze2022slurk}, which is a dialogue collection platform built to deal with server-side client events and API calls, ensuring participants could play the game online; additionally, it provides a straightforward and customizable logging system, as well as an off-the-shelf front-end interface with a built-in chat box feature.

All the images are ``cartoonish'' illustrations of real rooms and objects, in order to facilitate natural-language communication while creating a ``game'' feeling. There were a total of five movable objects: a pillow, pair of pants, trash bag, flat cap, and cowboy hat.
We found five items to strike a good balance between rich interactions and efficient gameplay.
Our game implementation prevented placing objects on top of each other in order to enforce nontrivial reference to locations through background landmarks.

The players were scored jointly, based on the mean Manhattan distance between identical objects. 
The closer the two common objects were placed on the grid, i.e., the smaller the distance between them was, the higher the score the pair received. 
The score was normalized on a scale from 0 to 100, contributing to the typical game "feel". Participants with very high scores (>99) got awarded a bonus. 

%% file: strategies.tex

\section{Game playing strategies}
\label{sec:strats}

We gathered a dataset of 73 games by crowd-sourcing participants via Prolific. We used this data to analyze human dialogue behavior in a collaborative environment.

\subsection{Collaboration strategies}

The participants in our dataset exhibited a number of distinct collaboration strategies, manually detected based on the players' contribution to the task-solving process. Examples of each strategy can be found in Appendix \ref{app:strats}.
Crucially, what we call the ``Leader'' strategy -- in which one player always dominates the collaboration -- is a minority.

\textbf{Leader.} (32.9\%\footnote{The brackets contain the percentage of total games employing each strategy.})
    One party predominantly leads, the other predominantly follows. It includes different situations: the explicit case (the players outwardly decide who should give instructions), cases where one player imposes the leader role and the other accepts it, or those where one player has to prompt the other for placements. The leader may or may not remain consistent across the two rounds---a swap in leadership was observed in 25\% of all games, whereas 67\% of games had a consistent leader. The remaining 8\% were "miscommunication" cases, where both users attempt to maintain the leader role.

\textbf{Back and forth.} (37.0\%) 	
    Both parties participate actively in solving the task, and the problem solving load is split between the two players. It contains the explicit case (the parties decided to each present a new placement for alternating objects), and the more natural case (one party suggests a new placement, the other accepts and follows up with a suggestion for another object). 

\begin{figure*}
    \makebox[\textwidth][c]{\includegraphics[width=0.99\textwidth]{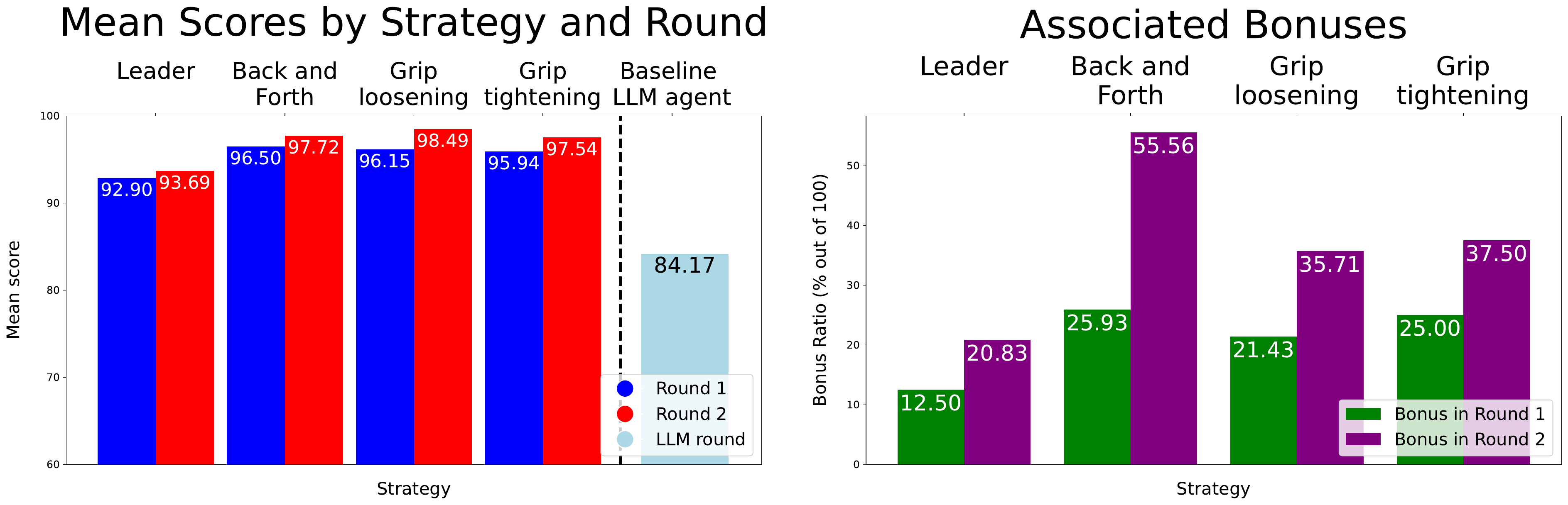}}
    \caption{Overview of strategies; \textbf{left} graph shows the mean scores in each round for each strategy (out of 100), while the \textbf{right} graph shows the distribution of bonuses (score > 99) per strategy in each round (expressed in \%).}
    \label{fig:strategies-overview}
\end{figure*}

 \textbf{Grip Tightening.} (11.0\%)
     The players move from a back-and-forth to a leader strategy. 
	Present either in cases where the first round does not go as smoothly as expected (resulting in one user taking the leader role onto themselves), or when the players have established a successful task-solving approach in the first round which can be carried out sufficiently well and more efficiently by only one player in the subsequent round.

 \textbf{Grip Loosening.} (19.2\%)		
    The players move from a leader to a back-and-forth strategy. 
    The first round typically contains a user that did not fully understand the task or was reluctant to communicate, resulting in the other player having to take the initiative and lead the game. The initially reluctant user would catch on by the end of the first round, and be more willing and ready to engage in a back and forth in the second round.

\subsection{Dominance scores}

Subsequently, we calculate a \emph{dominance score} for each player in each round of a game, capturing the extent to which one player dominates the way in which gameplay decisions are made. We assign a high dominance score to a player with high verbosity (mean message length) and high volume (percentage of messages sent, out of 100).

\newcommand{\vol}{\mathsf{volume}}
\newcommand{\dominance}{{\mathcal D}}
\newcommand{\verbo}{\mathsf{verbosity}}

More specifically, let A be the player with the higher volume and B the other player. We let $RD = (\vol_A - \vol_B) / (\vol_A + \vol_B)$ be the relative volume advantage of player A. Then we define
$$
\begin{array}{rcl}
\dominance_A &= &\verbo_A \cdot L(RD)\\
\dominance_B &= &\verbo_B \cdot (1 - L(RD)),
\end{array}$$

\noindent
where $L(x) = 1/(1+e^{-x})$ is the logistic function, so as to dampen large differences and emphasise smaller ones.
 \begin{table}[b]
 	\centering
 	\begin{tabular}{l|cc}
 		\multicolumn{1}{c|}{\textbf{Strategy}} & \multicolumn{1}{c}{\textbf{Round 1}} & \multicolumn{1}{c}{\textbf{Round 2}} \\ \hline
 		\textbf{leader} & 1.468 & 2.374 \\ \hline
 		\textbf{back and forth} & 1.17 & 0.981 \\ \hline
 		\textbf{grip tightening} & 0.884 & 1.696 \\ \hline
 		\textbf{grip loosening} & 1.421 & 0.988
 	\end{tabular}
 	\caption{Mean difference in the two players' dominance scores for each round (columns) in each strategy (rows).}
 	\label{tab:mean-dom-diff}
 \end{table}

We observe distinct patterns in each strategy's mean dominance score difference and its development across the two rounds (see Table \ref{tab:mean-dom-diff}), corresponding to their qualitative descriptions: in the leader case, one player has a much higher dominance score than the other in both rounds, whereas in the back and forth case, it is low across both rounds. In the grip tightening case, the dominance score difference is significantly higher in the second round than the first, indicating a change from a more balanced to an asymmetric approach, while the opposite is true in the grip loosening case.

\subsection{Impact of strategy on task success}

Figure \ref{fig:strategies-overview} breaks down game performance by strategy.
The figure on the left shows mean scores in each round for the four collaboration strategies; the figure on the right plots the proportion of games that received a bonus (score of 99 or more).
It is clear that the Leader strategy underperforms with respect to the others, with back-and-forth providing the greatest boost of bonus games from the first to the second round. This illustrates that our placement game is played most effectively by pairs who take a balanced approach to collaboration.

A key difference between our game an earlier reconstruction games is that our game forces the players to negotiate a goal state rather than being able to navigate to a predefined one. 
Moreover, the partial observability of the environment greatly impedes a leading player's ability to monitor the other player's actions and gauge the success of their leadership. Together, these features of our game seem to be sufficient to encourage balanced play.

%% file: modeling.tex
\section{Baseline LLM agent}
\label{sec:modeling}

Our game is intended as a testbed for computational agents that collaborate effectively with humans. To gauge how challenging it is for such agents, we evaluated a simple baseline agent based on LLMs.

The agent enforces a Leader strategy, with the human player as the leader, by asking the human player for instructions in the first message and remaining passive and reactive otherwise. It uses an LLM to perform simple semantic parsing of the human's instruction into triples of the form (object to move, landmark in the scene, spatial relation) and then uses simple handwritten rules to map such triples into $(x, y)$ positions in the scene. For instance, if the centerpoint of the fridge is at position $(x, y)$, the description ``above the fridge'' will be resolved to $(x, y-10)$. We use few-shot instruction giving with GPT 3.5 Turbo Instruct \cite{openai_gpt3}; see Appendix \ref{app:agent} for details.

In an online evaluation with ten human participants, the agent obtained a mean score of 84.17 (bottom left of Fig.~\ref{fig:strategies-overview}). This shows that the task is within reach of LLM-based agents; at the same time, the agent considerably lags behind even the human-human Leader strategy, suggesting that effective collaboration remains a challenge.

%% file: conclusion.tex
\section{Conclusion}

We have presented a 2D object placement game which is suitable for eliciting dialogues with varied collaboration strategies. This is in contrast to earlier dialogue games, in which one player typically takes the lead. The key innovation of our game is that players must negotiate their joint goal state. A baseline computational agent achieves a task performance that is within reach of, but still considerably below human performance, indicating that variants of our game would be an interesting and challenging platform for investigating human-computer collaboration.

In the future, it would be interesting to explore even more balanced versions of the game, e.g.\ by adding rules that increase the cost of failed collaboration. Another avenue of future research is to investigate the interplay of collaboration strategy and mutual adaptation of the player's lexica.

%% file: appendix.tex
\section{Appendix}
\label{app}

\subsection{Game environment design.}
\label{app:design}
\begin{figure*}[ht]
	\centering
    \includegraphics[width=1\linewidth]{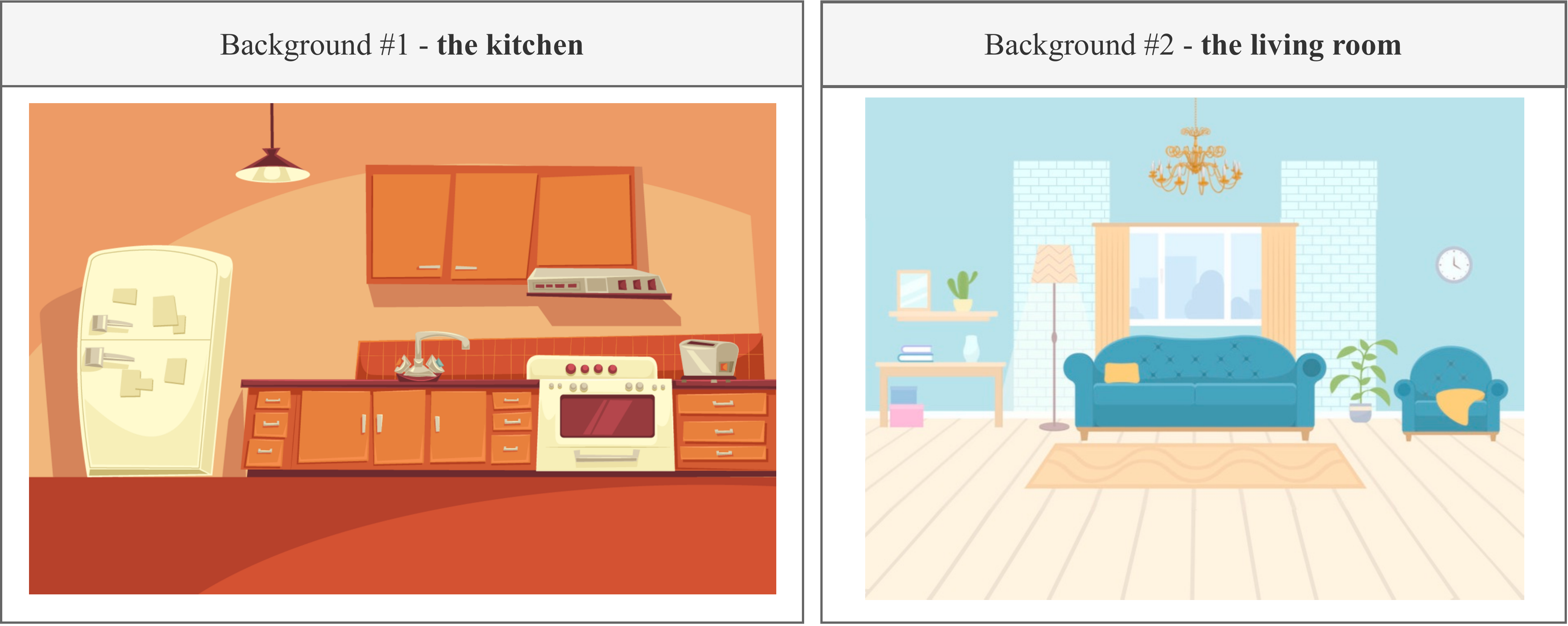}
	\caption{The background images for the two rounds.}
	\label{fig:bg}
\end{figure*}

Figure \ref{fig:bg} depicts the two background images used for the two rounds.

\subsection{Baseline agent.}
\label{app:agent}

\noindent \textbf{System logic.}    Each message that the human sends is analyzed by the agent following the steps from Table \ref{tab:sys-logic}. First, the agent determines if the user's message contained instructions, by using the input message together with the base (first prompt) and passing it to an LLM. If this step results in a \texttt{TRUE}, the system moves on to step 2, consisting of two extraction steps: in the first one, the  agent extracts the movable (target) object and static (reference) object, and in the second one, it extracts the placement direction of the target in reference to the static object. Table \ref{tab:allowed-terms} contains an overview of allowed terms for each extraction category. These entities are used in Step 3, which is a rule-based altering of the agent's world state, following the rule set from Table \ref{tab:constraint-rules} and hard-coded positions of the reference objects (this information corresponds to information available to the human, i.e., seeing one's own board). The next section of the appendix contains the base prompts.

\begin{table}
\begin{tabular}{|p{1.2cm}|p{5.6cm}|}
\hline
{\textbf{Step}} & \multicolumn{1}{c|}{\textbf{Description}} \\ 
\hline
\textbf{Step 1\dag} & \textbf{verify} if the message contains a set of instructions \\
\hline
\multirow{4}{*}{\textbf{Step 2\dag}}    & - \textbf{parse} the message\\& - for each group \footnotesize{(target, landmark, direction)}:\\
& 1. \textbf{extract} the term \\
& \,2. \textbf{map} the term to one of the predefined allowed terms \\ 
\hline
\textbf{Step 3$*$} & \textbf{change} the position of the objects according to Step 2 based on predefined constraints\\ 
\hline
\end{tabular}
\caption{A table showcasing the logic the baseline agent followed in order to complete the task. LLM-based steps are labeled with \dag, whereas rule-based ones have a $*$.}
\label{tab:sys-logic}
\end{table}
\begin{table}
	\centering
	\begin{tabular}{|c|c|c|}
		{\textbf{target}} & {\textbf{landmark}} & {\textbf{direction}} \\ \hline
		pillow & fridge  & on \\ 
		cowboy & toaster & next to \\ 
		cap & lamp & above \\ 
		pants & oven  & below \\ 
		garbage & stove & \\ 
		& counter & \\ 
		& sink & \\
	\end{tabular}
	\caption{All allowed terms per group; the extracted objects from the message are mapped to one term from each list.}
	\label{tab:allowed-terms}
\end{table}
\begin{table}
	\centering
	\begin{tabular}{c|c|c|}
		\cline{2-3}
		\multicolumn{1}{l|}{}                  & \multicolumn{1}{l|}{\textbf{new \textit{x[t]}}} & \multicolumn{1}{l|}{\textbf{new\textit{ y[t]}}} \\ \hline
		\multicolumn{1}{|c|}{\textbf{on}}      & x[r]                                   & y[r]                                   \\ \hline
		\multicolumn{1}{|c|}{\textbf{next to}} & x[r] + 10                              & y[r]                                   \\ \hline
		\multicolumn{1}{|c|}{\textbf{above}}   & x[r]                                   & y[r] - 10                              \\ \hline
		\multicolumn{1}{|c|}{\textbf{below}}   & x[r]                                   & y[r] + 10                              \\ \hline
	\end{tabular}
	\caption{The movement constraints for position manipulation. The first column contains the directions; the second and third columns refer to the target object (t)'s new x and y coordinates with respect to the reference landmark (r).}
	\label{tab:constraint-rules}
\end{table}

\newpage
\noindent \textbf{Prompts.} 	Here we provide the prompts we used for the LLM part of the agent.

\textbf{1.} The base of the prompt used to extract the placement location, in reference to a static object.
\begin{verbatim}
'''you are playing a game with another 
player in which you have to follow 
their instructions about where to put 
certain objects. i will give you a 
message and i want you to tell me if 
it contains a set of instructions. 
don't provide explanation, just give
me the output (True or False).
examples:
[user 1]: place the lamp on the fridge
[you]: True

[user 1]: can you put the knife in the 
drawer?
[you]: True

[user 1]: do you have a toaster?
[you]: False

[user 1]: what objects do you have?
[you]: False

[user 1']: let's place the pan on top of 
the lamp
[you]: True

[user 1]: put hat on sink
[you]: True

[user 1]: lamp on toilet
[you]: True''' 
\end{verbatim}
\textbf{2.}	The base of the prompt used to extract the static (reference) and movable (target) object.

\begin{verbatim}
'''i will give you a set of instruct-
ions and i want you to extract two 
things: one, the object that should 
be moved. then, i want you to compare 
it to the following four words and 
return the one it is most close to. 
the objects are: garbage, cowboy, 
cap, pants, pillow. next, i want you 
to extract the location where the 
object should be placed. then, match 
the output place with one of the 
possible places: fridge, counter, toast-
er, lamp, stove, oven, sink. don't 
provide explanation, just give me the 
output. for example: 
user 1: put the pillow to the right of 
the fridge
you: pillow, fridge

user 1: put the jeans on the stove
you: pants, stove

user 1: let's place the cushion on 
the ceiling light
you: pillow, lamp

user 1: place the garbagebag in 
the upper right corner of the counter
you: garbage, counter

user 1: cowboy hat to the left of 
the water faucet
you: cowboy, sink

user 1: the other hat on the right 
behind the pants
you: cap, toaster

user 1: garbage bag on top of 
lamp stand
you: garbage, lamp

user 1: let's place the blue hat 
on the toaster
you: cap, toaster

user 1: put peaky blinders hat 
in the oven
you: cap, oven'''
\end{verbatim}
\textbf{3.}	
The base of the prompt used to extract the placement location, in reference to a static object.

\begin{verbatim}
'''i will give you a set of instructions 
and i want you to extract the key spatial 
word or phrase.  then,  i want you to com-
pare it to the following four words and 
return the one it is most close to. the 
words are: above, below, next to, on. 
don't provide explanation, just give me 
the output. for example: 
[user 1]: put the knife to the right of 
the fridge
[you]: next to 

[user 1]: put the pan above the oven
[you]: above

[user 1]: place the toilet paper in the 
upper right corner of the cupboard
[you]: on

[user 1]: cowboy hat to the left of 
the water faucet
[you]: next to

[user 1]: the cowboy hat on the right 
behind the pants
[you]: next to 

[user 1]: pillow under the sink
[you]: below

[user 1]: garbage bag on top of lamp 
stand
[you]: above'''
\end{verbatim}

\subsection{Strategy examples}
\label{app:strats}
\begin{figure*}[ht]
	\centering
	\includegraphics[width=1\linewidth]{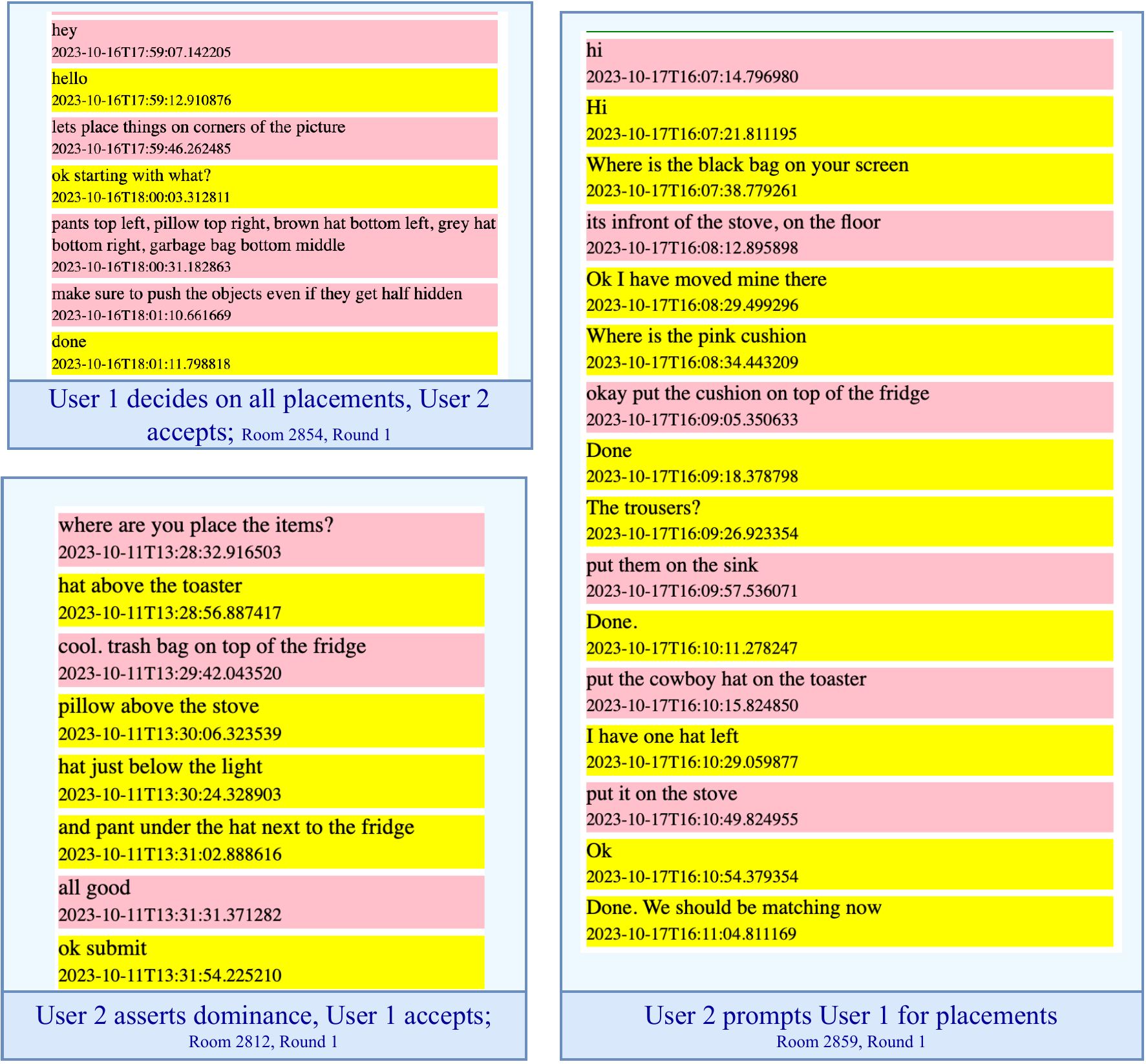}
	\caption{Leader strategy example}
	\label{fig:leader}
\end{figure*}

\begin{figure*}[ht]
	\centering
	\includegraphics[width=1\linewidth]{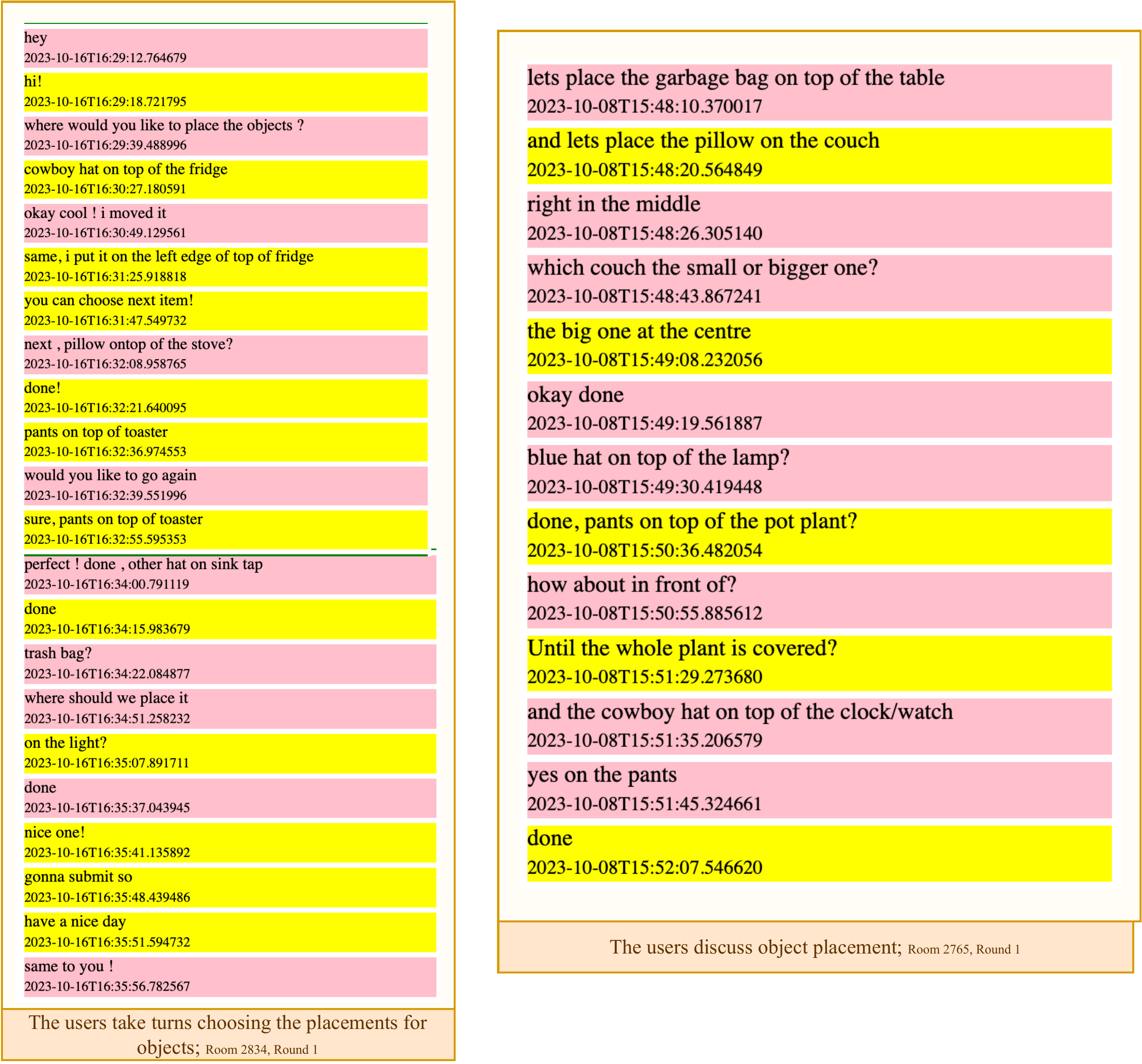}
	\caption{Back and forth strategy example}
	\label{fig:back-and-forth}
\end{figure*}

\begin{figure*}[ht]
	\centering
	\includegraphics[width=1\linewidth]{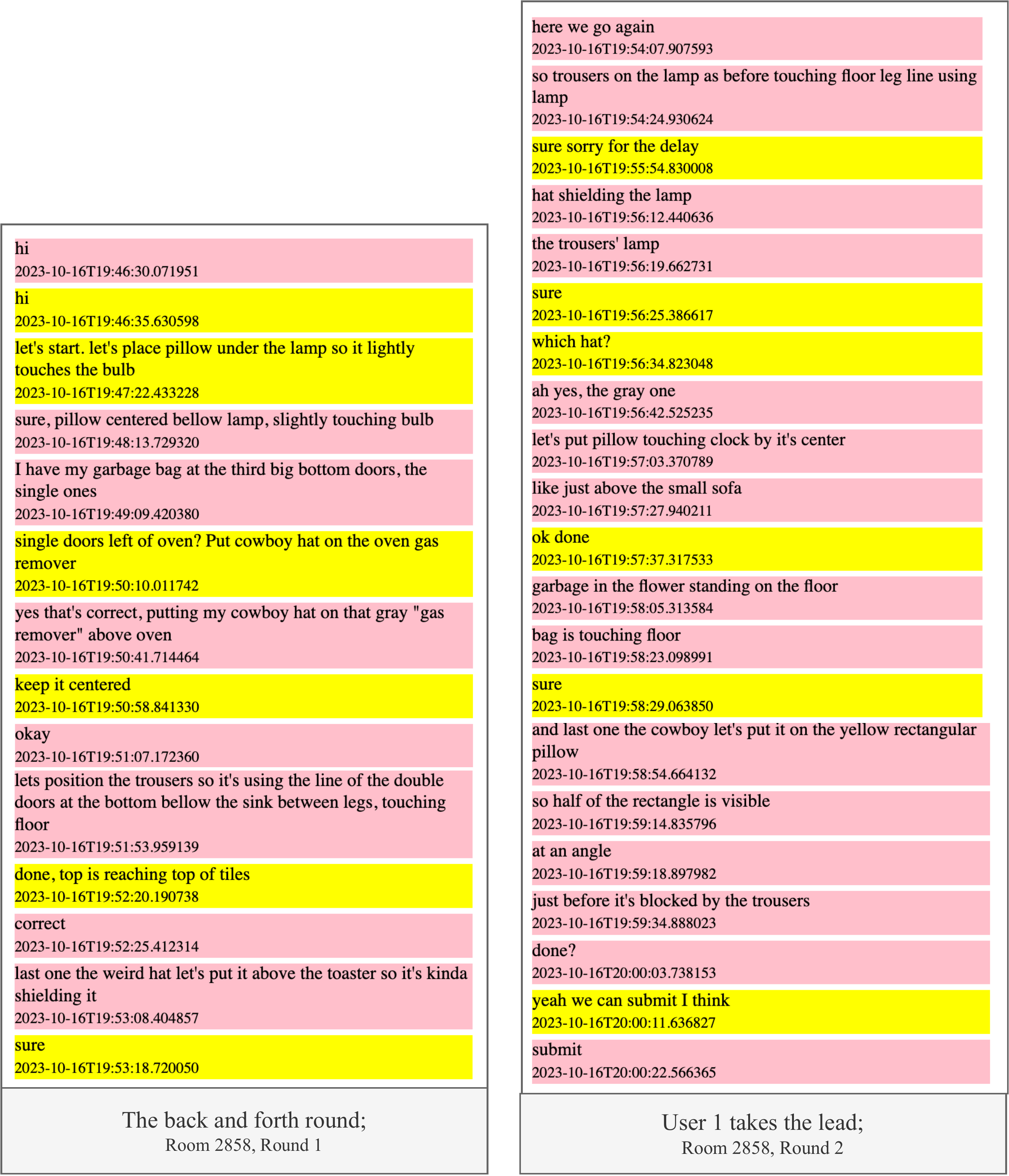}
	\caption{Grip tightening strategy example}
	\label{fig:grip-tighten}
\end{figure*}

\begin{figure*}[ht]
	\centering
	\includegraphics[width=1\linewidth]{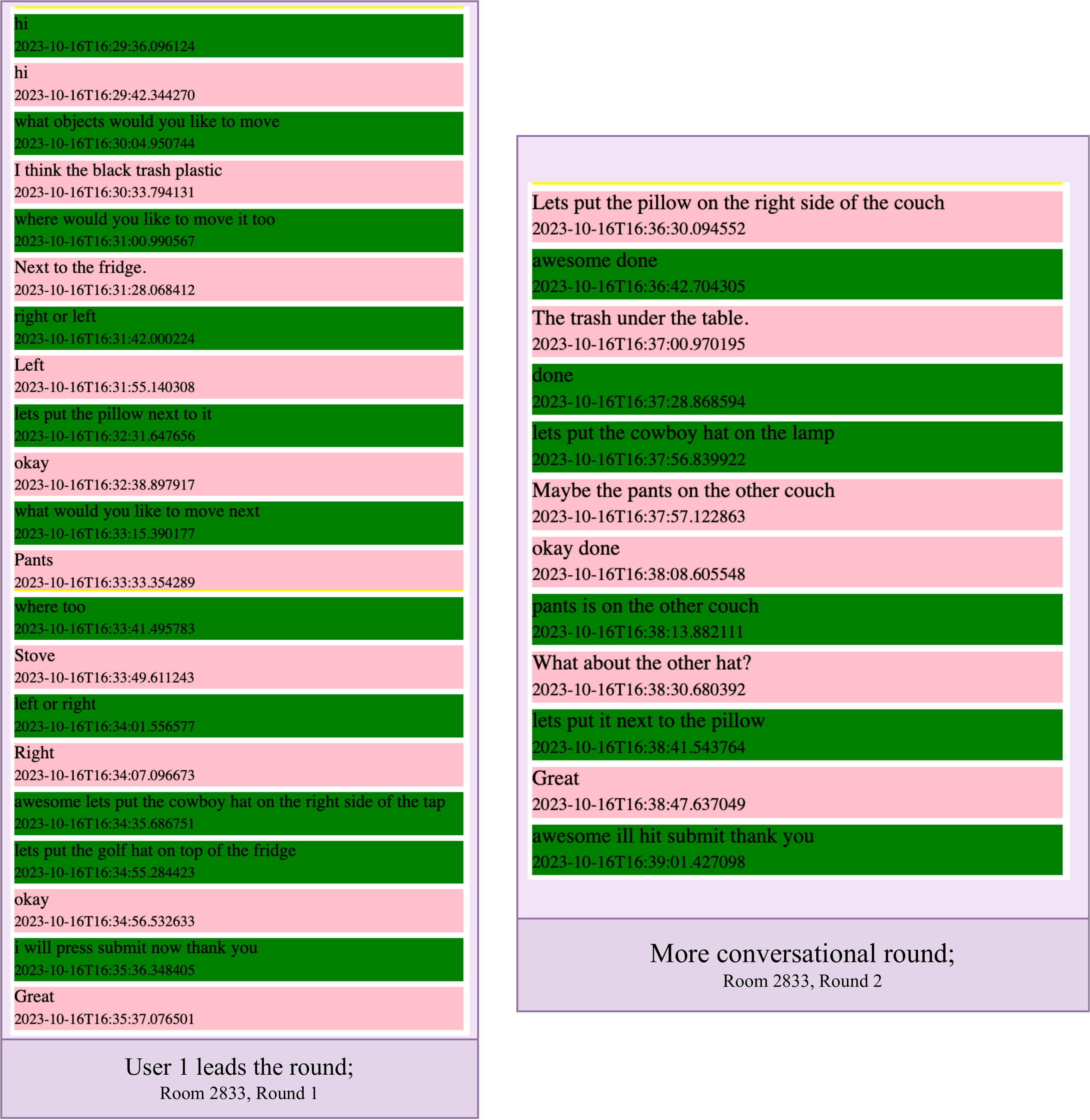}
	\caption{Grip loosening strategy example}
	\label{fig:grip-loosen}
\end{figure*}

Figures \ref{fig:leader} to \ref{fig:grip-loosen} illustrate examples of different strategies, namely:

\begin{itemize}
	\item leader --- Figure \ref{fig:leader}
	\item back and forth --- Figure \ref{fig:back-and-forth}
	\item grip tightening --- Figure \ref{fig:grip-tighten}
	\item grip loosening --- Figure \ref{fig:grip-loosen}
\end{itemize}

